\documentclass[runningheads]{llncs}
\usepackage[T1]{fontenc}

\usepackage{subcaption}
\usepackage{graphicx}
\usepackage{amsmath}
\usepackage{amssymb}
\usepackage{mathtools}
\usepackage{booktabs}
\usepackage{multirow}
\usepackage{tabularx}
\usepackage{wrapfig}
\usepackage{adjustbox}
\usepackage{color}
\usepackage{cite}
\usepackage{float}
\usepackage{tikz}
\usetikzlibrary {arrows.meta}

\usepackage{hyperref}
\usepackage{cleveref}

\title{Functional Tensor Decompositions for Physics-Informed Neural Networks}
\author{
    Sai Karthikeya Vemuri\inst{1,2}\orcidID{0009-0003-6272-8603}
    \and
    Tim Büchner\inst{1}\orcidID{0000-0002-6879-552X}
    \and
    Julia Niebling\inst{2}\orcidID{0000-0001-5413-2234}
    \and
    Joachim Denzler\inst{1}\orcidID{0000-0002-3193-3300}
}
\authorrunning{S. Vemuri et al.}

\institute{
    Computer Vision Group, Friedrich Schiller University Jena, 07743 Jena, Germany
    \and
    Institute of Data Science, German Aerospace Center, 07745 Jena, Germany
    \email{sai.karthikeya.vemuri@uni-jena.de}
}
\begin{document}
\maketitle
\begin{abstract}
Physics-Informed Neural Networks (PINNs) have shown continuous and increasing promise in approximating partial differential equations (PDEs), although they remain constrained by the curse of dimensionality.
In this paper, we propose a generalized PINN version of the classical variable separable method.
To do this, we first show that, using the universal approximation theorem, a multivariate function can be approximated by the outer product of neural networks, whose inputs are separated variables.   
We leverage tensor decomposition forms to separate the variables in a PINN  setting. 
By employing Canonic Polyadic (CP), Tensor-Train (TT), and Tucker decomposition forms within the PINN framework, we create robust architectures for learning multivariate functions from separate neural networks connected by outer products. 
Our methodology significantly enhances the performance of PINNs, as evidenced by improved results on complex high-dimensional PDEs, including the 3d Helmholtz and 5d Poisson equations, among others.
This research underscores the potential of tensor decomposition-based variably separated PINNs to surpass the state-of-the-art, offering a compelling solution to the dimensionality challenge in PDE approximation.

\end{abstract}

\section{Introduction}

Employing existing physical knowledge within data-driven systems is important for scientists at the intersection of science, engineering, and machine/deep learning to enforce proper behavior within a model.
Physics-informed neural networks (PINNs)~\cite{Raissi2017} provide the paradigm to formulate rules of physics within the network architecture, such that the model learns from data and the underlying physics.
Consequently, PINNs gained traction within the scientific machine-learning community.
Mainly, PINNs are extensively applied to solve forward and inverse problems involving systems of differential equations.
They are applied in various areas of science ranging from Geophysics, Structural mechanics, and Fluid dynamics to Epidemiology~\cite{Berkhahn2022,vemuri,Cuomo2022, soilsai}. 
We refer the readers to the excellent overview of PINNs by Cuomo et al.~\cite{Cuomo2022}.

Like other numerical methods for solving partial differential equations (PDE), PINNs also suffer from the curse of dimensionality.
We need $n$ collocation points to solve a PDE accurately.
In that case, the solution space explodes as $n^d$ with each dimension $d$, and classical PINNs get into computational problems as neural networks struggle to resolve relevant features, often leading to erroneous solutions.
Such problems and failure modes are discussed in detail in~\cite{krishnapriyan2021characterizing, Maddu_2022,WANG2022110768, Wangntk}.

A classic method of solving differential equations is the variable separable method~\cite{raisinghaniaordinary}, where the solution is defined as the product of univariate functions.
This approach is limited to a few classes of differential equations.   
We propose a new PINN alternative to the classical variable separable method, which can solve many arbitrary multi-dimensional PDEs irrespective of the presence of a separable form.
We leverage tensor decomposition approaches and their functional forms and use individual neural networks to learn information along each dimension. 

There is an increasing interest in using multiple neural networks and their blending together to form more accurate and robust PINNs.
Moseley et al.~\cite{moseley2021finite} suggested dividing the domain of PDE solution into smaller subdomains and using individual neural networks to learn the solution within each subdomain. 
Haghighat et al.~\cite{HAGHIGHAT2021113741} propose an individual neural network for each variable in a mathematical model of solid mechanics consisting of five variables. 
Cai et al.~\cite{Cai2021}, by applying PINNs to the two-phase Stefan problem, use two neural networks to model the unknown interface between two different material phases and describe the phases' two temperature distributions~\cite{Lu2021, Lin_Maxey_Li_Karniadakis_2021}.
Jin et al.~\cite{JinMIO} used multiple neural networks to improve neural operators, which also come under the broad family of physics-informed machine learning, where a PDE solution operator is learned instead of solving a particular PDE.
They used multiple branch and trunk nets and combined them to train Deep Neural Operators efficiently.
Applying this to PINNs, Cho et al.~\cite{cho2023separable} introduced Separable PINNs, where they use separate neural networks per axis, thus reducing the number of collocation points, and they leverage forward mode automatic differentiation to decrease computation and memory costs significantly. 

Building upon these approaches, we introduce functional tensor decompositions as a generalized separation of variables method.
We leverage several tensor decomposition forms to separate the variables in a PINN setting and approximate each decomposition component using a neural network.
We show that these methods are more accurate and faster than state-of-the-art PINN architectures. 
Our four main contributions are as follows:

\begin{enumerate}
    \item 
        {
        We extend the classical variable separable methods with PINNs by leveraging functional tensor decomposition forms, where individual neural networks learn each component of the tensor decomposition.
        }
    \item
        {
        We extend the universal approximation theorem and show that any multivariate function can be approximated using the outer product of univariate neural networks, irrespective of whether a variable separable form exists.
        }
    \item 
        {
        We propose to use three functional tensor decomposition forms that can be combined into PINNs: Canonic-Polyadic (CP-PINN)\cite{cphitchcock}, Tensor-Train (TT-PINN)\cite{ttdecomp}, and Tucker decomposition\cite{Tucker1966}.
        }
    \item 
        {
        We demonstrate that our proposed method outperforms previous state-of-the-art PINN architectures for high-dimensional PDEs and requires fewer collocation points~\cite{cho2023separable}.
        Thus, it offers an effective means to mitigate the curse of dimensionality and provide a better representation of solutions.
        The code is available at \url{https://github.com/cvjena/TensorDecompositions4PINNs}
        }
\end{enumerate}

\section{Theoretical Background}
The classical variable separable method involves writing a multivariate function as a sum of products of univariate functions.
Such forms only exist for a limited number of functions/PDEs~\cite{raisinghaniaordinary}. 
When we decompose a multivariate function and learn each univariate function using a neural network, it is crucial to demonstrate that this approach is practical even when a separable form does not exist.
This ensures the method's general applicability to all types of PDEs.
To support this argument, we show in the following sections that a multi-dimensional function can be approximated using the outer products of neural networks, where a neural network represents each dimension with a sufficient rank.
After that, we explain in detail how tensor decompositions are utilized in PINNs and how such architectures could mitigate the curse of dimensionality and improve the speed and accuracy of PINNs.

\subsection{Universal Approximation Theorem}
In this section, we revisit the classic Universal Approximation Theorem~\cite{HORNIK1989359, Cybenko1989} and extend it to separable functions.
We show empirically that any continuous multivariate function $f \colon \mathbb{K} \rightarrow \mathbb{R}$ within a compact bounded d-dimensional set $\mathbb{K} \in R^d$ can be approximated by the outer product of $d$ neural networks.
Each neural network is a function of a single variable  $x_i (1 \leq i \leq d)$. 
Furthermore, we demonstrate how tensor decomposition forms can separate dimensions and how these components support solving PDEs using PINNs.

The Universal Approximation Theorem states that a feed-forward neural network with a single hidden layer can theoretically approximate any continuous function on a bounded domain with arbitrary accuracy~\cite{HORNIK1989359, Cybenko1989}. 
Hence, given a continuous function $f: \mathbb{R}^{d} \rightarrow \mathbb{R}$ and for any $\epsilon>0$, there exists a feed-forward neural network $\widehat{f}$, such that
\begin{equation}
    \left \| f(x) - \widehat{f}(x) \right \| < \epsilon ; \forall x \in \mathbb{K}.
\end{equation}
considering that for all $x$ inside of $\mathbb{K}$, which is a compact subset in $\mathbb{R}^{d}$, the inequality holds true.

Functional approximation problems, including those tackled by PINNs, involve high-dimensional spaces where the \emph{curse of dimensionality} becomes a significant issue.
Thus, significant challenges are posed in computation, the number of parameters needed, and the calculation of derivatives (an additional special case for PINNs).

A classical approach to address this \emph{curse of dimensionality} is the sum of separable functions and is based on reconstructing a multivariate function as a product of univariate functions~\cite{cho2023separable,raisinghaniaordinary,phdthesis}:
A $d$-variate function $f:\mathbb{K}\rightarrow \mathbb{R}$ can be written as
\begin{equation}
    f(x_1, x_2, \dots, x_d) = \sum_{j=1}^{r}\bigotimes_{i=1}^{d}g_{i}^{j}(x_i).
    \label{eq:sum_sep}
\end{equation}

Where $g_{i}^{j}$ are univariate functions, $j$ denotes the separation rank, representing the number of terms in the function $g_i$. 
We designate the operator $\bigotimes$ as the tensor product of vector spaces defined by individual univariate functions.
One of the important features of these functions is that they are not restricted from coming from a particular basis set. 
This is the exact point we would like to emphasize.
Since they are not restricted to being unique, we propose that neural networks can approximate these functions. 

We simplify the notation by rewriting the \Cref{eq:sum_sep} to omit separation rank, and the separated functions $g_{i}$ contain $r$ components corresponding to the $i$-th dimension.
This is written as:
\begin{equation}
    f(x_1,x_2....,x_d) = \bigotimes_{i=1}^{d}g_{i}(x_i).
    \label{eq:sep}
\end{equation}

Now, we use $d$ neural networks, e.g., per dimension of the problem, to approximate the individual functions $g_{i}$.
Now the tensor product of these univariate approximations is the approximation of $f$, denoted by $\widehat{f}$, as
\begin{equation}
    \widehat{f}(x_1,x_2....,x_d) = \bigotimes_{i=1}^{d}\widehat{g}_{i}(x_i,\theta_i).
    \label{eq:sep_nns}
\end{equation}

Where $\theta_i$ represents a neural network's trainable parameters (weights and biases). 
Following the above-mentioned \textit{Universal Approximation Theorem}, we can approximate a neural network $\widehat{g}_{i}$ for every individual component $g_{i}$~\cite{HORNIK1989359, Cybenko1989} and every $\epsilon_i\in\mathbb{R}$, such that 
\begin{equation}
    \left \| g_{i}(x_i) - \widehat{g}_{i}(x_i) \right \| < \epsilon_i ; \forall x_i \in \mathbb{K}_i \wedge 1 \leq i \leq d,
    \label{eq:5}
\end{equation}

where $\mathbb{K}_i$ is a compact subset of $\mathbb{R}$.
The error in approximating $f$ by $\widehat{f}$ can be now written as 
\begin{equation}
    \left \| f(x_1, x_2, \ldots, x_d) - \hat{f}(x_1, x_2, \ldots, x_d) \right \| = \left \| \bigotimes_{i=1}^{d}g_{i}(x_i) -  \bigotimes_{i=1}^{d}\widehat{g}_{i}(x_i,\theta_i) \right \|.
    \label{eq:6}
\end{equation}

Under certain reasonable assumptions that the univariate functional spaces are Banach in nature, and the norm is a reasonable cross-norm~\cite{phdthesis}, the norm of the outer product can be written as simply the norm of products.
This property is illustrated as follows: 
\begin{equation}
    \| \bigotimes_{i=1}^d x_i \| =  \prod_{i=1}^d \| x_i \|.
\end{equation}

Using this property and identity of difference of products, which is derived using mathematical induction in~\cite{phdthesis}, we expand the right-hand side of \Cref{eq:6}:
\begin{equation}
    \bigotimes_{i=1}^d g_{i} - \bigotimes_{i=1}^d \widehat{g}_{i} = \sum_{j=1}^d \left( \prod_{k=1}^{j-1} \widehat{g}_k \right) (g_j - \widehat{g}_j) \left( \prod_{l=j+1}^d g_{l} \right).
\end{equation}

Taking the norm and including the above equation in \Cref{eq:6}
\begin{equation}
    \| f - \widehat{f} \| = \left\| \sum_{j=1}^d \left( \prod_{k=1}^{j-1} \widehat{g}_k \right) (g_j - \widehat{g}_j) \left( \prod_{l=j+1}^d g_l \right) \right\|.
\end{equation}

Using the triangle inequality~\cite{rudin1964principles}, we obtain 
\begin{equation}
    \| f - \widehat{f} \| \leq \sum_{j=1}^d \left\| \left( \prod_{k=1}^{j-1} \widehat{g}_k \right) (g_j - \widehat{g}_j) \left( \prod_{l=j+1}^d g_l \right) \right\|.
\end{equation}

Finally, assuming the norm is sub-multiplicative, something like $L$, we get 
\begin{equation}
\| f - \widehat{f} \| \leq \sum_{j=1}^d \left( \prod_{k=1}^{j-1} \| \widehat{g}_k \| \right) \| g_j - \widehat{g}_j \| \left( \prod_{l=j+1}^d \| g_l \| \right).
\end{equation}

Simplified expression by using \Cref{eq:5}
\begin{equation}
    \| f - \widehat{f} \| \leq \sum_{j=1}^d \left( \prod_{k=1}^{j-1} \| \widehat{g}_k \| \right) \epsilon_j \left( \prod_{l=j+1}^d \| g_l \| \right).
\end{equation}

By appropriately choosing the approximation errors $\epsilon_j$ for each univariate function and making sure that the norms don't explode (i.e., weights and gradients do not explode)\cite{glorot10a,kingma2014adam}, we can ensure that the total error is less than any desired $\epsilon$.
Thus, the constructed multivariate approximation using outer products of univariate functions with large enough rank is also a universal function approximator. 
This shows that, theoretically, we can represent any arbitrary multivariate function, regardless of the existence of variable separable form, as the outer product of neural networks.
The inputs to these individual neural networks correspond to particular dimensions. 
The underlying ideas of triangle inequality and identity of differences are drawn from well-established theories in the field~\cite{phdthesis,cho2023separable, JinMIO}.

\subsection{Physics-Informed Neural Networks}
Physics-Informed Neural Networks (PINNs) are a class of neural networks that incorporate physical laws described by partial differential equations (PDEs) into the training process~\cite{Raissi2017}.
Unlike traditional neural networks that rely exclusively on data-driven learning, PINNs utilize the underlying model structure, e.g., the actual gradients in the loss function, to constrain the solution space effectively.
Thereby, prior knowledge and physically consistent constraints are enforced into the learning process.
Specifically, in a PINN, the loss function $\mathcal{L}$ is augmented with terms that enforce the PDE constraints.
Consider a PDE of the form:
\begin{equation}
    \mathcal{F}(\mathbf{x}, u(\mathbf{x}), \nabla u(\mathbf{x}), \nabla^2 u(\mathbf{x}), \dots) = 0,
    \label{eq:pde}
\end{equation}
where $x \in \mathbb{R}^{d}$ represents the spatial and temporal variables, and $u(x)$ describes the solution.
The loss function $\mathcal{L}$ is composed of the data loss $\mathcal{L}_{\text{data}}$ and physics loss $\mathcal{L}_{\text{physics}}$.
$\mathcal{L}_{\text{data}}$ captures the error between the predicted solution $u(x;\theta)$ and the observed data.
Furthermore, $\mathcal{L}_{\text{physics}}$ constrains the solution space such that the model abides by the underlying governing physics at collocation points within the domain. 
We formally define both as
\begin{align}
    \mathcal{L}_{\text{data}} &= \frac{1}{N} \sum_{i=1}^{N} \left( u(\mathbf{x}_i; \theta) - u_i^{\text{data}} \right)^2, \text{ and}
    \label{eq:l_data}\\
    \mathcal{L}_{\text{physics}} &= \frac{1}{M} \sum_{j=1}^{M} \left( \mathcal{F}(\mathbf{x}_j, u(\mathbf{x}_j; \theta), \nabla u(\mathbf{x}_j; \theta), \nabla^2 u(\mathbf{x}_j; \theta), \dots) \right)^2.
    \label{eq:l_phy}
\end{align}

Where $N$ and $M$ are data and collocation points, respectively. 
The interplay between these two loss functions is controlled by parameter $\lambda$~\cite{vemuri,mcclenny2022selfadaptive, Wangntk}.
The combined multi-objective loss function is given as:
\begin{equation}
    \mathcal{L} = \mathcal{L}_{\text{data}} + \lambda \cdot \mathcal{L}_{\text{physics}}.
    \label{eq:loss}
\end{equation}

As stated, the collocation points refer to points inside the domain where the physics is obeyed.
Generally, these are sampled uniformly in the domain to ensure the neural network learns the domain space.
The curse of dimensionality manifests in PINNs as the number of collocations grows exponentially for every additional dimension.
This challenges PINNs on many fronts, making them computationally expensive, and approximating solutions becomes increasingly difficult.
Such failure modes are more explained in the works like~\cite{vemuri, Wangntk, WANG2022110768, krishnapriyan2021characterizing, Maddu_2022}.

PINNs can be easily seen as a special case of functional approximation, where, along with some samples, we give the underlying PDE residual(physics)  from which the underlying function (solution of PDE) \emph{u} needs to be approximated. 
We propose to represent the solution of a PINN as the outer product of univariate neural networks to separate variables. 
This can be seen as the PINN counterpart of the classic variable separable method.
Since we have shown that the outer product of neural networks with a sufficient rank can approximate a multivariate function, this works even for cases where classical variable separable form does not exist, making a generalized separation of the variable method. The advantages of this approach over classical PINNs are as follows:

\begin{enumerate}
    \item{
        Requirement of fewer collocation points.
        While a classical PINN requires $n^d$ collocation points to sample a $d$-dimensional domain, our approach needs only $n \cdot d$ points, effectively mitigating the curse of dimensionality.
    }
    \item{
        The solution of the PDE is expressed in variable separable form, irrespective of classical variable separable form.
    }
    \item{
        We employ individual neural networks per dimension, allowing better feature representation and avoiding potential local minima in complex problems.
    }
\end{enumerate}

\subsection{Functional Tensor Decompositions for PINNs}
\begin{figure}[t]
    \centering
    \begin{subfigure}[t]{0.29\textwidth}
        \centering
        \resizebox{\textwidth}{!}{%
            \newcommand{\Width}{2}
\newcommand{\Height}{2}
\newcommand{\Depth}{-1.414}

\begin{tikzpicture}[cross/.style={path picture={ 
      \draw[black]
    (path picture bounding box.south east) -- (path picture bounding box.north west) (path picture bounding box.south west) -- (path picture bounding box.north east);
    }}]
    
    




    \draw[draw, anchor=center, cross] (5.45, 1,\Depth/2) circle (0.25);
    \node[draw, minimum width=2.0cm, minimum height=1cm, rotate=0,  anchor=center, blue,  fill=blue!20,  opacity=1.0] (A1) at (4.0, 1.0,\Depth/2) {$A_1(x_1)$};
    \node[draw, minimum width=1.0cm, minimum height=2cm, rotate=0,  anchor=center, orange,fill=yellow!20,opacity=1.0] (A2) at (6.5, 1.0,\Depth/2) {$A_2(x_2)$};
    \node[draw, minimum width=1.8cm, minimum height=1cm, rotate=40, anchor=south, green, fill=green!20, opacity=1.0] (A3) at (6.0, 2.3,\Depth/2) {$A_3(x_3)$};

    \node[anchor=north] at (A1.south) {$n_1 \times R$};
    \node[anchor=north] at (A2.south) {$n_2 \times R$};
    \node[anchor=south,rotate=40] at (A3.north) {$n_3 \times R$};
\end{tikzpicture}
        }
        \caption{Canonic-Polyadic\cite{cphitchcock}}
        \label{fig:decomp_cp}
    \end{subfigure}%
    \hfill
    \begin{subfigure}[t]{0.36\textwidth}
        \centering
        \resizebox{\textwidth}{!}{%
            \newcommand{\Width}{2}
\newcommand{\Height}{2}
\newcommand{\Depth}{-1.414}

\begin{tikzpicture}[cross/.style={path picture={ 
      \draw[black]
    (path picture bounding box.south east) -- (path picture bounding box.north west) (path picture bounding box.south west) -- (path picture bounding box.north east);
    }}]
    
    




    \node[draw, minimum width=1.5cm, minimum height=1cm, rotate=0,  anchor=center, blue,  fill=blue!20,  opacity=1.0] (A1) at (4.0, 1.0,\Depth/2) {$A_1(x_1)$};

    \draw[draw, anchor=center, cross] (5.0, 1,\Depth/2) circle (0.15);

    \node[draw, minimum width=2.0cm, minimum height=0.8cm, rotate=0,  anchor=center, orange, fill=yellow!20,opacity=1.0] (A2) at (6.25, 1.0,\Depth/2) {$A_2(x_2)$};
    \draw[orange,fill=yellow!20,opacity=0.8] (6.25+1, 1.0-0.4, \Depth/2) -- (6.25+1, 1.0-0.4, \Depth/2+\Depth/4) -- (6.25+1, 1.0+0.4, \Depth/2+\Depth/4) -- (6.25+1, 1.0+0.4, \Depth/2) -- cycle;
    \draw[orange,fill=yellow!20,opacity=0.8] (6.25-1, 1.0+0.4, \Depth/2) -- (6.25+1, 1.0+0.4, \Depth/2) -- (6.25+1, 1.0+0.4, \Depth/2+\Depth/4) -- (6.25-1, 1.0+0.4, \Depth/2+\Depth/4) -- cycle;
    
     \draw[draw, anchor=center, cross] (7.6, 1,\Depth/2) circle (0.15);
    
    \node[draw, minimum width=0.8cm, minimum height=2cm, rotate=0,  anchor=center, green,fill=green!20,opacity=1.0] (A3) at (8.5, 1.0,\Depth/2) {$A_3(x_3)$};

    \node[anchor=north] at (A1.south) {$n_1 \times R_1$};
    \node[anchor=north] at (A2.south) {$R_1 \times n_2 \times R_2$};
    \node[anchor=north] at (A3.south) {$R_2 \times n_3$};

\end{tikzpicture}
        }
        \caption{Tensor Train~\cite{ttdecomp}}
        \label{fig:decomp_tt}
    \end{subfigure}
    \hfill
     \begin{subfigure}[t]{0.34\textwidth}
        \centering
         \resizebox{\textwidth}{!}{%
            \newcommand{\Width}{2}
\newcommand{\Height}{2}
\newcommand{\Depth}{-1.414}

\begin{tikzpicture}[cross/.style={path picture={ 
      \draw[black]
    (path picture bounding box.south east) -- (path picture bounding box.north west) (path picture bounding box.south west) -- (path picture bounding box.north east);
    }}]
    
    \node[draw, minimum width=1.5cm, minimum height=1cm, rotate=0,  anchor=center, blue,  fill=blue!20,  opacity=1.0] (A1) at (4.0, 1.0,\Depth/2) {$A_1(x_1)$};
    \draw[draw, anchor=center, cross] (5.0, 1,\Depth/2) circle (0.15);

    \coordinate (C1) at (5.25, 1.0+0.4, \Depth/2);
    \coordinate (C2) at (5.25, 1.0-0.4, \Depth/2);
    \coordinate (C3) at (6.0, 1.0-0.8, \Depth/2);
    \coordinate (C4) at (6.0, 1.0, \Depth/2);
    
    \coordinate (C5) at (6.75, 1.0-0.4, \Depth/2);
    \coordinate (C6) at (6.75, 1.0+0.4, \Depth/2);
    \coordinate (C7) at (6.0, 1.0+0.8, \Depth/2);
    
    \draw[violet,fill=violet!20,opacity=0.8] (C1) -- (C2) -- (C3) -- (C4) -- cycle; 
    \draw[violet,fill=violet!20,opacity=0.8] (C3) -- (C5) -- (C6) -- (C4) -- cycle; 
    \draw[violet,fill=violet!20,opacity=0.8] (C1) -- (C4) -- (C6) -- (C7) -- cycle; 

    \node[anchor=center] at (6, 1.0+0.4,\Depth/2) {$C$};

    \draw[draw, anchor=center, cross] (7.0, 1,\Depth/2) circle (0.15);
    \node[draw, minimum width=0.8cm, minimum height=2cm, rotate=0,  anchor=center, orange,fill=yellow!20,opacity=1.0] (A2) at (7.9, 1.0,\Depth/2) {$A_2(x_2)$};

    \draw[draw, anchor=center, cross] (6.0, 2.0,\Depth/2) circle (0.15);
    \node[draw, minimum width=1.8cm, minimum height=1cm, rotate=35, anchor=south, green, fill=green!20, opacity=1.0] (A3) at (6.7, 2.7,\Depth/2) {$A_3(x_3)$};

    \node[anchor=north] at (A1.south) {$n_1 \times R_1$};
    \node[anchor=north] at (C3.south) {$R_1 \times R_2 \times R_3$};
    \node[anchor=north] at (A2.south) {$n_2 \times R_2$};
    \node[anchor=south,rotate=35] at (A3.north) {$n_3 \times R_3$};

\end{tikzpicture}
        }
        \caption{Tucker~\cite{Tucker1966}}
        \label{fig:decomp_tu}
    \end{subfigure}
    \caption{
        We provide a schematic visualization for the tensor decompositions (\subref{fig:decomp_cp})-(\subref{fig:decomp_tu}) on the examples for $d=3$.
        The shape of the factor tensors ($A$) is written on the bottom of each component.
        Tucker~\cite{Tucker1966} additionally has one core tensor $C$.
    }
    \label{fig:TensorDecompositions}
\end{figure}
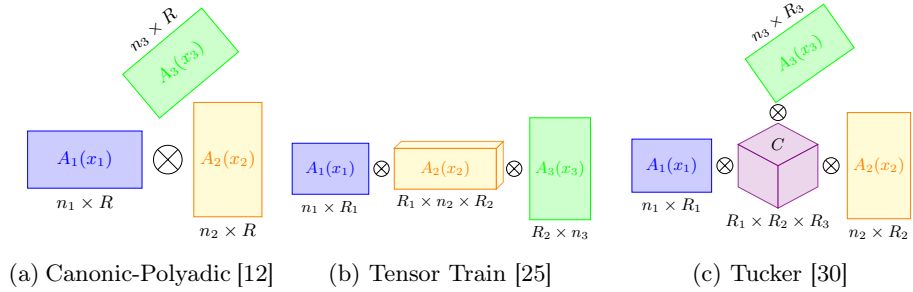

We leverage tensor decomposition forms to achieve the separation of variables.
This approach decomposes a high-dimensional tensor into smaller components by approximating multivariate functions using outer products of univariate functions.
We refer to this technique as \emph{functional tensor decomposition}.
A separate neural network is responsible for learning each component in the tensor decomposition.
This work discusses three tensor decomposition forms: Canonic-Polyadic~\cite{cphitchcock}, Tensor-Train~\cite{ttdecomp}, and Tucker decompositions~\cite{Tucker1966}.
We provide definitions and schematics, and later the inclusion into the PINN architectures, but recommend the work of~\cite{TensorDecompsReview} for a broad overview of tensor decomposition.

\subsubsection{Canonic-Polyadic Decomposition}
(CP) involves decomposing a $d$ order tensor into $d$ factor matrices of a specified rank $R$~\cite{cphitchcock} similar to Separable PINN~\cite{cho2023separable} as shown in \Cref{fig:decomp_cp}.  
Mathematically, for a multi-dimensional tensor $f$, the CP decomposition is written as

\begin{equation}
f(x_1, x_2, \dotsc, x_d) \approx [[A_1(x_1), A_2(x_2), \dotsc, A_d(x_d)]],
\end{equation}
 where $[[\cdot]]$ denotes tensor product operation with $A_1 \in \mathbb{R}^{n_1 \times R},\ldots, A_d\in \mathbb{R}^{n_d \times R}$ being factor matrices, for $n_i$ points along each $i$-th dimension.

\begin{figure}[t]
    \centering
    \begin{subfigure}[t]{0.3\textwidth}
        \centering
        \resizebox{\textwidth}{!}{%
            \newcommand{\Width}{2}
\newcommand{\Height}{2}
\newcommand{\Depth}{-1.414}

\begin{tikzpicture}[cross/.style={path picture={ 
      \draw[black]
    (path picture bounding box.south east) -- (path picture bounding box.north west) (path picture bounding box.south west) -- (path picture bounding box.north east);
    }}]

    \node[draw, minimum width=1.5cm, minimum height=0.3cm, rotate=0,  anchor=center, blue,  fill=blue!5,   opacity=0.8] (i1) at (4.2, 7.5) {\scriptsize $n_1 \times 1$};
    \node[draw, minimum width=1.5cm, minimum height=0.3cm, rotate=0,  anchor=center, orange,fill=yellow!5, opacity=0.8] (i2) at (6.0, 7.5) {\scriptsize $n_2 \times 1$};
    \node[draw, minimum width=1.5cm, minimum height=0.3cm, rotate=0,  anchor=center, green, fill=green!5,  opacity=0.8] (i3) at (7.8, 7.5) {\scriptsize $n_3 \times 1$};

    \node[draw, minimum width=1.5cm, minimum height=1.5cm, rotate=0,  anchor=center, blue,  fill=blue!5,   opacity=0.8] (NN1) at (4.2, 6.0) {\scriptsize $\text{NN}_1$};
    \node[draw, minimum width=1.5cm, minimum height=1.5cm, rotate=0,  anchor=center, orange,fill=yellow!5, opacity=0.8] (NN2) at (6.0, 6.0) {\scriptsize $\text{NN}_2$};
    \node[draw, minimum width=1.5cm, minimum height=1.5cm, rotate=0,  anchor=center, green, fill=green!5,  opacity=0.8] (NN3) at (7.8, 6.0) {\scriptsize $\text{NN}_3$};

    \node[draw, minimum width=1.3cm, minimum height=0.8cm, rotate=0,  anchor=center, blue,  fill=blue!20,  opacity=1.0] (A1) at (4.2, 4.0) {\scriptsize $A_1(x_1)$};
    \node[draw, minimum width=0.5cm, minimum height=1.3cm, rotate=0,  anchor=center, orange,fill=yellow!20,opacity=1.0] (A2) at (6.0, 4.0) {\scriptsize $A_2(x_2)$};
    \node[draw, minimum width=1.3cm, minimum height=0.7cm, rotate=0,  anchor=center, green, fill=green!20, opacity=1.0] (A3) at (7.8, 4.0) {\scriptsize $A_3(x_3)$};

    \draw[-{Stealth[length=2mm]}] (i1.south) -- (NN1.north);
    \draw[-{Stealth[length=2mm]}] (i2.south) -- (NN2.north);
    \draw[-{Stealth[length=2mm]}] (i3.south) -- (NN3.north);

    \draw[-{Stealth[length=2mm]}] (NN1.south) -- (A1.north);
    \draw[-{Stealth[length=2mm]}] (NN2.south) -- (A2.north);
    \draw[-{Stealth[length=2mm]}] (NN3.south) -- (A3.north);

    \draw[-{Stealth[length=2mm]}] (A1.south) -- (4.2,2.4) -- (6.0-0.25, 2.4);
    \draw[-{Stealth[length=2mm]}] (A2.south) -- (6.0, 2.4+0.25);
    \draw[-{Stealth[length=2mm]}] (A3.south) -- (7.8, 2.4) -- (6.0+0.25, 2.4);
    
    \draw[draw, anchor=center, cross] (6.0, 2.4) circle (0.25);
    \draw[-{Stealth[length=2mm]}] (6.0, 2.4-0.25) -- (6.0, 1.8);

    \node[anchor=south] at (i1.north) {$x_1$};
    \node[anchor=south] at (i2.north) {$x_2$};
    \node[anchor=south] at (i3.north) {$x_3$};

    \node[anchor=south,rotate=90] at (A1.west) {\scriptsize $n_1 \times R$};
    \node[anchor=south,rotate=90] at (A2.west) {\scriptsize $n_2 \times R$};
    \node[anchor=south,rotate=90] at (A3.west) {\scriptsize $n_3 \times R$};

    \coordinate (C1) at (5.25, 1.4);
    \coordinate (C2) at (5.25, 0.6);
    \coordinate (C3) at (6.00, 0.2);
    \coordinate (C4) at (6.00, 1.0);
    
    \coordinate (C5) at (6.75, 0.6);
    \coordinate (C6) at (6.75, 1.4);
    \coordinate (C7) at (6.00, 1.8);
    
    \draw[orange,fill=orange!20,opacity=0.8] (C1) -- (C2) -- (C3) -- (C4) -- cycle; 
    \draw[orange,fill=orange!20,opacity=0.8] (C3) -- (C5) -- (C6) -- (C4) -- cycle; 
    \draw[orange,fill=orange!20,opacity=0.8] (C1) -- (C4) -- (C6) -- (C7) -- cycle; 

    \node[anchor=south,rotate=90] at (5.25, 1.0) {\scriptsize $n_1 \times n_2 \times n_3$}; 
    \node[anchor=north,rotate=0] at (C3.south) {$f(x_1,x_2,x_3)$};

\end{tikzpicture}
        }
        \caption{CP-PINN}
        \label{fig:cppinn}
    \end{subfigure}
    \hfill
    \begin{subfigure}[t]{0.3\textwidth}
        \centering
        \resizebox{\textwidth}{!}{%
            \newcommand{\Width}{2}
\newcommand{\Height}{2}
\newcommand{\Depth}{-1.414}

\begin{tikzpicture}[cross/.style={path picture={ 
      \draw[black]
    (path picture bounding box.south east) -- (path picture bounding box.north west) (path picture bounding box.south west) -- (path picture bounding box.north east);
    }}]

    \node[draw, minimum width=1.5cm, minimum height=0.3cm, rotate=0,  anchor=center, blue,  fill=blue!5,   opacity=0.8] (i1) at (4.2, 7.5) {\scriptsize $n_1 \times 1$};
    \node[draw, minimum width=1.5cm, minimum height=0.3cm, rotate=0,  anchor=center, orange,fill=yellow!5, opacity=0.8] (i2) at (6.0, 7.5) {\scriptsize $n_2 \times 1$};
    \node[draw, minimum width=1.5cm, minimum height=0.3cm, rotate=0,  anchor=center, green, fill=green!5,  opacity=0.8] (i3) at (7.8, 7.5) {\scriptsize $n_3 \times 1$};

    \node[draw, minimum width=1.5cm, minimum height=1.5cm, rotate=0,  anchor=center, blue,  fill=blue!5,   opacity=0.8] (NN1) at (4.2, 6.0) {\scriptsize $\text{NN}_1$};
    \node[draw, minimum width=1.5cm, minimum height=1.5cm, rotate=0,  anchor=center, orange,fill=yellow!5, opacity=0.8] (NN2) at (6.0, 6.0) {\scriptsize $\text{NN}_2$};
    \node[draw, minimum width=1.5cm, minimum height=1.5cm, rotate=0,  anchor=center, green, fill=green!5,  opacity=0.8] (NN3) at (7.8, 6.0) {\scriptsize $\text{NN}_3$};

    \node[draw, minimum width=1.3cm, minimum height=0.8cm, rotate=0,  anchor=center, blue,  fill=blue!20,  opacity=1.0] (A1) at (4.2, 4.0) {\scriptsize $A_1(x_1)$};
    \node[draw, minimum width=0.5cm, minimum height=1.3cm, rotate=0,  anchor=center, orange,fill=yellow!20,opacity=1.0] (A2) at (6.0, 4.0) {\scriptsize $A_2(x_2)$};
    \node[draw, minimum width=1.3cm, minimum height=0.7cm, rotate=0,  anchor=center, green, fill=green!20, opacity=1.0] (A3) at (7.8, 4.0) {\scriptsize $A_3(x_3)$};

    \draw[orange,fill=yellow!20,opacity=1.0] (5.465,4.0+0.65) -- (6.545,4.0+0.65) -- (6.745,4.0+0.65+0.2) -- (5.745,4.0+0.65+0.2) -- cycle;
    \draw[orange,fill=yellow!20,opacity=1.0] (6.545,4.0+0.65) -- (6.545,4.0-0.65) -- (6.745,4.0-0.65+0.2) -- (6.745,4.0+0.65+0.2)   -- cycle;

    \draw[-{Stealth[length=2mm]}] (i1.south) -- (NN1.north);
    \draw[-{Stealth[length=2mm]}] (i2.south) -- (NN2.north);
    \draw[-{Stealth[length=2mm]}] (i3.south) -- (NN3.north);

    \draw[-{Stealth[length=2mm]}] (NN1.south) -- (A1.north);
    \draw[-{Stealth[length=2mm]}] (NN2.south) -- (A2.north);
    \draw[-{Stealth[length=2mm]}] (NN3.south) -- (A3.north);

    \draw[-{Stealth[length=2mm]}] (A1.south) -- (4.2,2.4) -- (6.0-0.25, 2.4);
    \draw[-{Stealth[length=2mm]}] (A2.south) -- (6.0, 2.4+0.25);
    \draw[-{Stealth[length=2mm]}] (A3.south) -- (7.8, 2.4) -- (6.0+0.25, 2.4);
    
    \draw[draw, anchor=center, cross] (6.0, 2.4) circle (0.25);
    \draw[-{Stealth[length=2mm]}] (6.0, 2.4-0.25) -- (6.0, 1.8);

    \node[anchor=south] at (i1.north) {$x_1$};
    \node[anchor=south] at (i2.north) {$x_2$};
    \node[anchor=south] at (i3.north) {$x_3$};

    \node[anchor=south,rotate=90] at (A1.west) {\scriptsize $n_1 \times R_1$};
    \node[anchor=south,rotate=90] at (A2.west) {\scriptsize $R_1 \times n_2 \times R_2$};
    \node[anchor=south,rotate=90] at (A3.west) {\scriptsize $R_2 \times n_3$};

    \coordinate (C1) at (5.25, 1.4);
    \coordinate (C2) at (5.25, 0.6);
    \coordinate (C3) at (6.00, 0.2);
    \coordinate (C4) at (6.00, 1.0);
    
    \coordinate (C5) at (6.75, 0.6);
    \coordinate (C6) at (6.75, 1.4);
    \coordinate (C7) at (6.00, 1.8);
    
    \draw[orange,fill=orange!20,opacity=0.8] (C1) -- (C2) -- (C3) -- (C4) -- cycle; 
    \draw[orange,fill=orange!20,opacity=0.8] (C3) -- (C5) -- (C6) -- (C4) -- cycle; 
    \draw[orange,fill=orange!20,opacity=0.8] (C1) -- (C4) -- (C6) -- (C7) -- cycle; 

    \node[anchor=south,rotate=90] at (5.25, 1.0) {\scriptsize $n_1 \times n_2 \times n_3$}; 
    \node[anchor=north,rotate=0] at (C3.south) {$f(x_1,x_2,x_3)$};

\end{tikzpicture}
        }
        \caption{TT-PINN}
        \label{fig:ttpinn}
    \end{subfigure}
    \hfill
     \begin{subfigure}[t]{0.3\textwidth}
        \centering
        \resizebox{\textwidth}{!}{%
            \newcommand{\Width}{2}
\newcommand{\Height}{2}
\newcommand{\Depth}{-1.414}

\begin{tikzpicture}[cross/.style={path picture={ 
      \draw[black]
    (path picture bounding box.south east) -- (path picture bounding box.north west) (path picture bounding box.south west) -- (path picture bounding box.north east);
    }}]

    \node[draw, minimum width=1.5cm, minimum height=0.3cm, rotate=0,  anchor=center, blue,  fill=blue!5,   opacity=0.8] (i1) at (4.2, 7.5) {\scriptsize $n_1 \times 1$};
    \node[draw, minimum width=1.5cm, minimum height=0.3cm, rotate=0,  anchor=center, orange,fill=yellow!5, opacity=0.8] (i2) at (6.0, 7.5) {\scriptsize $n_2 \times 1$};
    \node[draw, minimum width=1.5cm, minimum height=0.3cm, rotate=0,  anchor=center, green, fill=green!5,  opacity=0.8] (i3) at (7.8, 7.5) {\scriptsize $n_3 \times 1$};

    \node[draw, minimum width=1.5cm, minimum height=1.5cm, rotate=0,  anchor=center, blue,  fill=blue!5,   opacity=0.8] (NN1) at (4.2, 6.0) {\scriptsize $\text{NN}_1$};
    \node[draw, minimum width=1.5cm, minimum height=1.5cm, rotate=0,  anchor=center, orange,fill=yellow!5, opacity=0.8] (NN2) at (6.0, 6.0) {\scriptsize $\text{NN}_2$};
    \node[draw, minimum width=1.5cm, minimum height=1.5cm, rotate=0,  anchor=center, green, fill=green!5,  opacity=0.8] (NN3) at (7.8, 6.0) {\scriptsize $\text{NN}_3$};

    \node[draw, minimum width=1.3cm, minimum height=0.8cm, rotate=0,  anchor=center, blue,  fill=blue!20,  opacity=1.0] (A1) at (4.2, 4.0) {\scriptsize $A_1(x_1)$};
    \node[draw, minimum width=0.5cm, minimum height=1.3cm, rotate=0,  anchor=center, orange,fill=yellow!20,opacity=1.0] (A2) at (6.0, 4.0) {\scriptsize $A_2(x_2)$};
    \node[draw, minimum width=1.3cm, minimum height=0.7cm, rotate=0,  anchor=center, green, fill=green!20, opacity=1.0] (A3) at (7.8, 4.0) {\scriptsize $A_3(x_3)$};

    \coordinate (P1) at (6.0-0.25, 2.5-0.00);
    \coordinate (P2) at (6.0-0.25, 2.5-0.25);
    \coordinate (P3) at (6.0-0.00, 2.5-0.50);
    \coordinate (P4) at (6.0-0.00, 2.5-0.25);
    \coordinate (P5) at (6.0+0.25, 2.5-0.25);
    \coordinate (P6) at (6.0+0.25, 2.5-0.00);
    \coordinate (P7) at (6.0-0.00, 2.5+0.25);
    
    \draw[violet,fill=violet!20,opacity=0.8] (P1) -- (P2) -- (P3) -- (P4) -- cycle; 
    \draw[violet,fill=violet!20,opacity=0.8] (P3) -- (P5) -- (P6) -- (P4) -- cycle; 
    \draw[violet,fill=violet!20,opacity=0.8] (P1) -- (P4) -- (P6) -- (P7) -- cycle; 
    
    \coordinate (C1) at (5.25, 1.4);
    \coordinate (C2) at (5.25, 0.6);
    \coordinate (C3) at (6.00, 0.2);
    \coordinate (C4) at (6.00, 1.0);
    
    \coordinate (C5) at (6.75, 0.6);
    \coordinate (C6) at (6.75, 1.4);
    \coordinate (C7) at (6.00, 1.8);
    
    \draw[orange,fill=orange!20,opacity=0.8] (C1) -- (C2) -- (C3) -- (C4) -- cycle; 
    \draw[orange,fill=orange!20,opacity=0.8] (C3) -- (C5) -- (C6) -- (C4) -- cycle; 
    \draw[orange,fill=orange!20,opacity=0.8] (C1) -- (C4) -- (C6) -- (C7) -- cycle; 

    \draw[-{Stealth[length=2mm]}] (i1.south) -- (NN1.north);
    \draw[-{Stealth[length=2mm]}] (i2.south) -- (NN2.north);
    \draw[-{Stealth[length=2mm]}] (i3.south) -- (NN3.north);

    \draw[-{Stealth[length=2mm]}] (NN1.south) -- (A1.north);
    \draw[-{Stealth[length=2mm]}] (NN2.south) -- (A2.north);
    \draw[-{Stealth[length=2mm]}] (NN3.south) -- (A3.north);

    \draw[-{Stealth[length=2mm]}] (A1.south) -- (4.2,2.4) -- (6.0-0.25, 2.4);
    \draw[-{Stealth[length=2mm]}] (A2.south) -- (P7);
    \draw[-{Stealth[length=2mm]}] (A3.south) -- (7.8, 2.4) -- (6.0+0.25, 2.4);
    
    \node[anchor=south] at (i1.north) {$x_1$};
    \node[anchor=south] at (i2.north) {$x_2$};
    \node[anchor=south] at (i3.north) {$x_3$};

    \node[anchor=north] at (6.0-0.00, 2.5+0.20) {\scriptsize $C$};

    \node[anchor=south,rotate=90] at (A1.west) {\scriptsize $n_1 \times R_1$};
    \node[anchor=south,rotate=90] at (A2.west) {\scriptsize $n_2 \times R_2$};
    \node[anchor=south,rotate=90] at (A3.west) {\scriptsize $n_3 \times R_3$};

    \node[anchor=north west] at (P5.south) {\tiny $R_1 \times R_2 \times R_3$};

    \draw[-{Stealth[length=2mm]}] (P3.south) -- (C7.north);

    \node[anchor=south,rotate=90] at (5.25, 1.0) {\scriptsize $n_1 \times n_2 \times n_3$}; 
    \node[anchor=north,rotate=0] at (C3.south) {$f(x_1,x_2,x_3)$};

\end{tikzpicture}
        }
        \caption{Tucker-PINN}
        \label{fig:tupinn}
    \end{subfigure}
    \caption{
        Functional tensor decomposition forms within the PINN model architecture:
        The approximation of each component matrix based on a single variable is done with an individual neural network.
        These outputs are then combined as in the Canonic-Polyadic~\cite{cphitchcock} (\subref{fig:cppinn}), Tensor-Train~\cite{ttdecomp} (\subref{fig:ttpinn}) or Tucker~\cite{Tucker1966} (\subref{fig:tupinn}) manner.
    }
    \label{fig:TensorPINNS}
\end{figure}
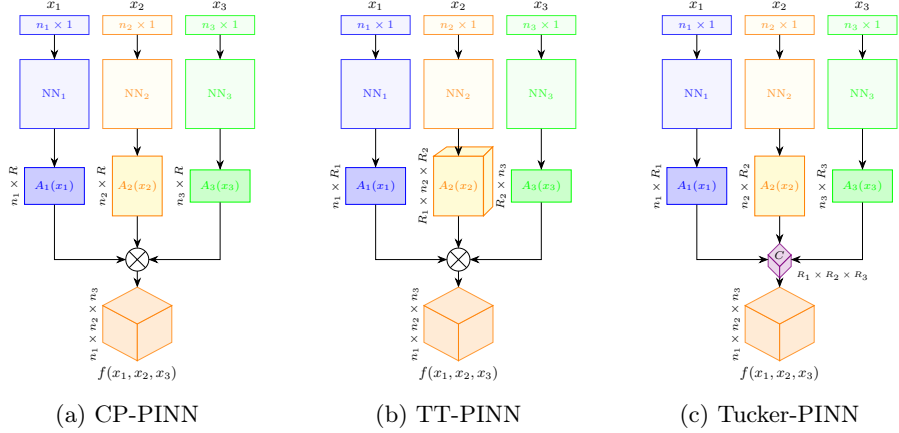

\subsubsection{Tensor-Train Decomposition}
(TT) represents a high-dimensional tensor as a sequence of low-dimensional tensors (cores) connected in a chain (train)~\cite{ttdecomp}, with an example shown in \Cref{fig:decomp_tt}.
Similar to the matrix notation CP, we have 
\begin{equation}
f(x_1, x_2, \ldots, x_d) \approx [[A_1(x_1), A_2(x_2), \ldots, A_d(x_d)]]
\end{equation}
where each core tensor $A_i \in \mathbb{R}^{R_{i-1}\times n_i \times R_i}$, and $R_0 = R_d = 1$, where $n_i$ is number of the points in the $i$-th dimension. Unlike CP, TT connects tensors belonging to the adjacent dimensions, making it more stable\cite{ttdecomp}. 

\subsubsection{Tucker Decomposition}
generalizes CP by decomposing a tensor into a core tensor multiplied by matrices along each mode~\cite{Tucker1966}, visualized in \Cref{fig:decomp_tu}.  
Therefore, by updating the CP matrix notation, we obtain 
\begin{equation}
f(x_1, x_2, \ldots, x_d) \approx [[\mathcal{C}; A_1(x_1), A_2(x_2), \ldots, A_d(x_d)]]~.
\end{equation}
We denote $\mathcal{C} \in \mathbb{R}^{R_1 \times R_2 \times R_3 \times \ldots \times R_d}$ as the core tensor and  $A_1 \in \mathbb{R}^{n_1 \times R_1}, A_2 \in \mathbb{R}^{n_2 \times R_2}, \dotsc, A_d \in \mathbb{R}^{n_d \times R_d}$ are factor matrices.
Unlike both CP and TT decomposition, the core connects all the dimensions, making it even more stable, with more parameters~\cite{Tucker1966}.

\subsubsection{Functional tensor decomposition forms in PINNs}
We now use the aforementioned tensor decomposition in a PINN setup.
As described earlier, we assume the solution of a PDE that needs to be solved by a PINN is decomposed into components constituting any of the tensor decomposition mentioned above, and a neural network learns each component. 
Therefore, we propose three architectures based on functional tensor decompositions: CP-PINN, TT-PINN, and Tucker-PINN. The schematics given in \Cref{fig:TensorDecompositions} and \Cref{fig:TensorPINNS} are for a three-dimensional PDE, but the concepts scales to arbitrary dimensions $(\geq 3)$.
For CP-PINN and Tucker-PINN, each network outputs a factor matrix of shape $n \times R$, where $n$ still denotes the input dimension and $R$ the desired rank of the decomposition.
For this paper's scope, we consider that the ranks for all components of Tucker-PINN and TT-PINN are set to the same integer $R$.  
Additionally, for Tucker-PINN, we initialize the core tensor as an orthogonal and trainable parameter to learn the entries during training.
For TT-PINN, each network outputs either the train start or end part with the shape $n \times r$ or the train middle of shape $r \times n \times r$.

\section{Experiments}
We solve benchmark PDEs in three dimensions and more to demonstrate the effectiveness of our tensor decomposition architectures CP-PINN, TT-PINN, and Tucker-PINN.
We compare our results with the original PINN architecture~\cite{Raissi2017} and other state-of-the-art PINN models~\cite{LuLu2019,mcclenny2022selfadaptive,wang2022respecting}. 
Each variable is put into a four-layer feed-forward neural network with $\tanh(\cdot)$ activation functions for the proposed PINN architecture.
The feature depth per layer corresponds to the rank unless specified otherwise.
A network's input is collocation points along a single dimension of shape $n \times 1$, with $n$ being the number of available points.

Our \emph{functional tensor decomposition} models, the PDE simulation code, and experiments are created in JAX~\cite{jax2018github}.
We adopt the implementation of forward gradients from~\cite{cho2023separable}.
The overall setup adheres to the conventional PINN framework, comprising a composite loss function (Equation \ref{eq:loss}) that combines data and PDE residual terms with no weighting, i.e., by setting $\lambda=1$. 
All models are trained using Adam optimizer~\cite{kingma2014adam} with learning rate $1e^{-3}$ and for $50000$ iterations.
The performance metric is the $L^2$ error between predicted and simulated solutions.
The tests are conducted on an NVIDIA GeForce GTX 1080 GPU, with reported relative speeds in iterations per second (IT/s).

First, we choose two three-dimensional PDE benchmarks, (2+1)d Klein-Gordon and 3d Helmholtz equation~\cite{lu2021deepxde,takamoto2023pdebench,cho2023separable,zeng2024feature}, for investigating the performance of our \emph{functional tensor decomposition} based PINNs.
Problems of this high dimension are computationally challenging for PINNs yet frequently arise in real-world applications.
The boundary/initial conditions and visualizations are reported in \Cref{tab:list_of_pdes} upper half.
We compare our models against state-of-the-art methods like gradient-based PINN~\cite{Raissi2017}, G-PINN~\cite{Yu_2022}, SA-PINN~\cite{mcclenny2022selfadaptive}, and Causal-PINN~\cite{wang2022respecting} (all implemented via PINA~\cite{coscia2023physics}).
We omit SPINN~\cite{cho2023separable} due to the same nature as CP-PINN.
We ensured the solution was converged for all the benchmarks, and the training setup was as close as given in the original source.
We experiment with multiple collocation points and ranks to evaluate the influence of these hyperparameters on the general model architecture.

\begin{table}[H]
    \centering
    \caption{
        We provide an overview of the four PDE experiment setups that were used to investigate the capabilities of our tensor decomposition PINNs.
        The plots are best viewed digitally and in color.
    }
    \begin{tabular}{p{0.49\textwidth}p{0.49\textwidth}}
    \toprule
        \multicolumn{1}{c}{(2+1)d Klein-Gordon~\cite{cho2023separable}} & \multicolumn{1}{c}{3d Helmholtz~\cite{takamoto2023pdebench}}\\
        \midrule
        \adjustbox{valign=t}{
            $\!\begin{aligned}
                \label{eq:kg}
                    f &= \partial_{tt} u - \Delta u + u^2,   \\ 
                    u &= \left(x+y\right) \cdot \cos(2t), \\
                    u(x,y,0) &= x+y + x \cdot y \cdot \sin(2t), \\
                    [x,y] &\in [-1,1]^2, t \in [0,10]
            \end{aligned}$
        }
        & 
        \adjustbox{valign=t}{
            $\!\begin{aligned}
                \Delta u + k^2u &= q , x \in [-1,1]^3,\\
                u(x) &= 0, x \in \partial \Omega, \\
                   q &= - (a_1\pi)^{2}u-(a_2\pi)^{2}u\\
                     &~~~~-(a_3\pi)^{2}u+k^{2}u,\\
                   u &= \prod_{i=1}^{3} \sin(a_i \pi x_i)
                \label{eq:helm}
            \end{aligned}$
        }
        \\
        \multicolumn{1}{c}{
            \includegraphics[width=0.35\textwidth]{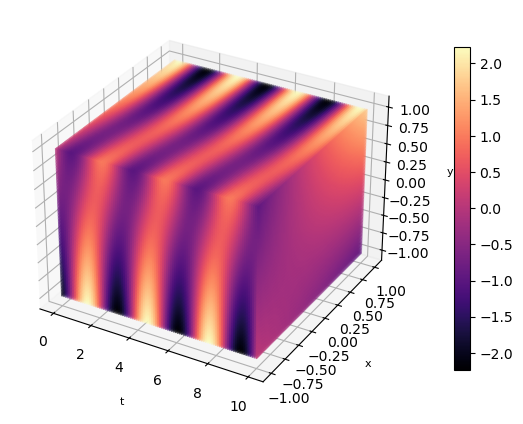}
        }
        &
        \multicolumn{1}{c}{
            \includegraphics[width=0.35\textwidth]{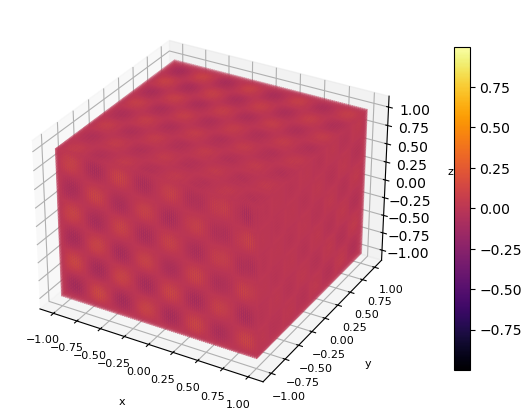}
        }
        \\
        \toprule
        \multicolumn{1}{c}{(2+1)d Flow mixing\cite{CHIU2022114909}} & \multicolumn{1}{c}{5d Poisson\cite{zeng2024feature}}\\ 
        \midrule
        \adjustbox{valign=t}{
            $\!\begin{aligned}
            0 &= \frac{\partial u}{\partial t} + a\frac{\partial u}{\partial x} + b\frac{\partial u}{\partial y},\\
            a(x, y) &= -\frac{v_t \cdot y}{v_{\text{t,max}} \cdot r},\\
            b(x, y) &=  \frac{v_t \cdot x}{v_{\text{t,max}} \cdot r},\\
            v_t &= \frac{\tanh(r)}{\tanh^2(r)},\\
            r &= \sqrt{x^2 + y^2},\\
            u &= -\tanh(\frac{y}{2}\cos(\frac{v_t}{r \cdot v_{t,\text{max}}} t)\\
              &~~~~- \frac{x}{2}\sin(\frac{v_t}{r\cdot v_{t,\text{max}}} t))\\
            \text{with } t &\in [0, 4], x \in [-4, 4], y \in [-4, 4],\\
            \text{and }& v_{\text{t,max}} = 0.385\\
            \end{aligned}$
        }
        &
        \adjustbox{valign=t}{
            $\!\begin{aligned}
                \Delta u &= -  \frac{\pi^2}{4} \sum_{i = 1}^{n} \sin(\frac{\pi}{2}x_i),\\
                       u &= \sum_{i=1}^{n} \left ( \sin \left ( \frac{\pi}{2}x_i \right ) \right ),\\
                       \text{with } x &\in \left [ 0, 1 \right ] , n=5. 
           \end{aligned}$
        }
        \\
        \multicolumn{1}{c}{
            \includegraphics[width=0.35\textwidth]{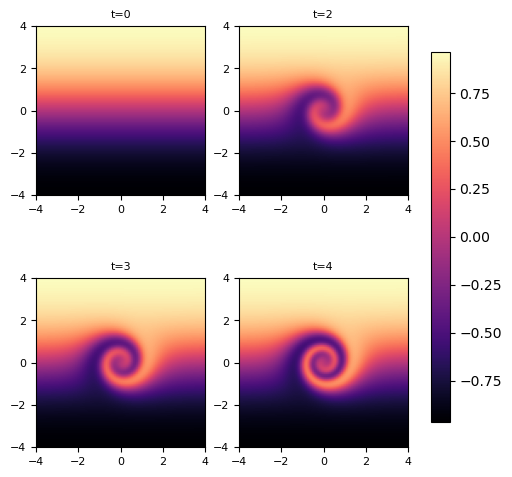}
        }
        &
        \multicolumn{1}{c}{
            \includegraphics[width=0.35\textwidth]{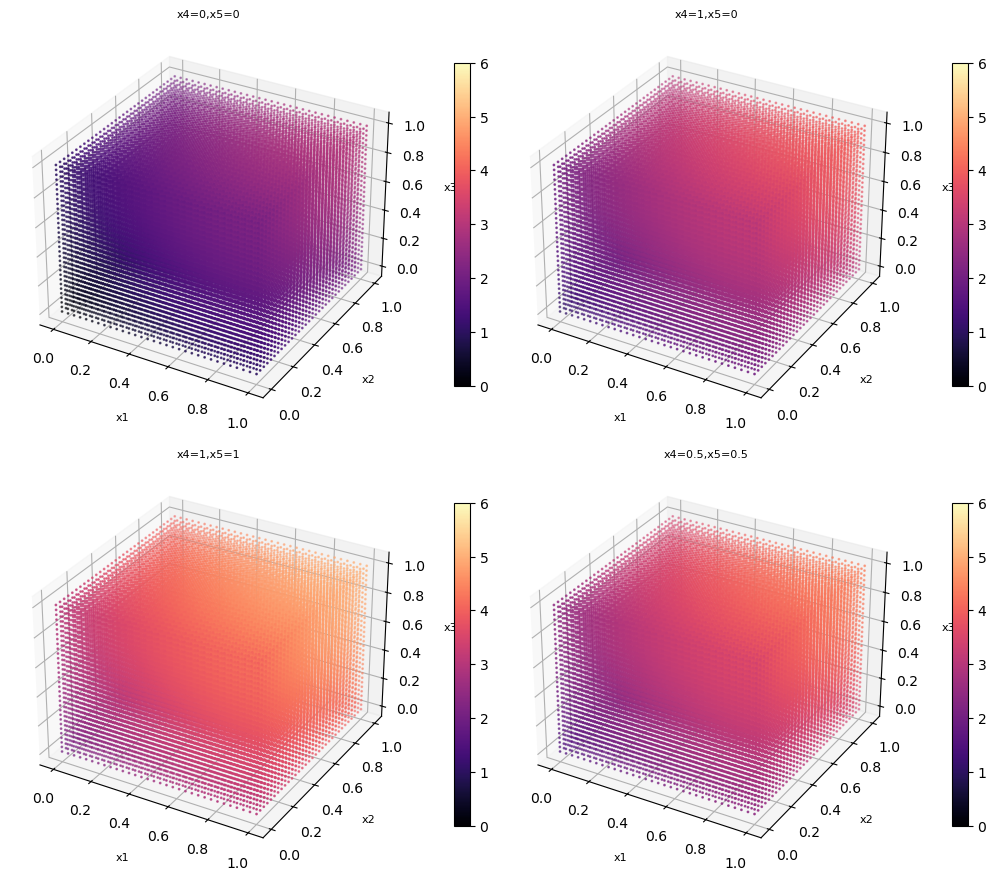}
        }
        \\
        \bottomrule
    \end{tabular}
    \label{tab:list_of_pdes}
\end{table}

\begin{table}[t]
    \centering
    \caption{
        (2+1)d Klein Gordon Equation and 3d Helmholtz are solved by various architectures.
        The general trend is that a larger rank leads and a larger number of points leads to a better solution.
        Furthermore, we observe that our tensor decomposition PINNs are a magnitude of 10 better at solving these problems. 
        The best performance per task is \underline{\underline{double underlined}}, and the second best is \underline{underlined}.
    }
    \newcolumntype{e}{>{\centering}m{1.45cm}}
\newcolumntype{a}{>{\centering}m{2.5cm}}
\newcolumntype{d}{>{\centering\arraybackslash}m{2.2cm}}
\begin{tabular}{leeead}
    \toprule
    PINN model & Rank & Points & $\mathcal{L}^{2}$ Klein-Gordon $\downarrow$ & {$\mathcal{L}^{2}$ Helmholtz} $\downarrow$ & Speed (IT/s) $\uparrow$ \\
    \midrule
    Vanilla PINN\cite{Raissi2017}                 & -    & $32^{3}$        & 0.092                   &           0.998    & 20      \\
    G-PINN\cite{Lu2021}                       & -    & $32^{3}$        & 0.073                   &           0.794    & 16        \\
    SA-PINN\cite{mcclenny2022selfadaptive}                      & -    & $32^{3}$        & 0.095                   &           0.920    & 15        \\
    Causal-PINN\cite{wang2022respecting}                  & -    & $32^{3}$        & 0.041                   &           0.406    & 3        \\
    \midrule
    \multirow{5}{*}{CP-PINN}     & 8    & $64\times3$        & 0.050                   &          0.061  & 364           \\
                                 & 16   & $64\times3$        & 0.025                   &          0.051  & 347           \\
                                 & 32   & $16\times3$        & 0.069                   &          0.063  & 357           \\
                                 & 32   & $32\times3$        & 0.055                   &          0.060  & 343           \\
                                 & 32   & $64\times3$        & \underline{\underline{0.008}}       &          \underline{\underline{0.040}} & 327                  \\
    \midrule
    \multirow{5}{*}{TT-PINN}     & 8    & $64\times3$        & 0.068                   &          0.064  & 358           \\
                                 & 16   & $64\times3$        & 0.043                   &          0.061  & 353           \\
                                 & 32   & $16\times3$        & 0.088                   &          0.060  & 356           \\
                                 & 32   & $32\times3$        & 0.079                   &          0.055  & 310           \\
                                 & 32   & $64\times3$        & \underline{0.010} &    \underline{0.048}  & 305           \\
    \midrule
    \multirow{5}{*}{Tucker-PINN} & 8    & $64\times3$        & 0.061                   &          0.079  & 345           \\
                                 & 16   & $64\times3$        & 0.053                   &          0.076  & 328           \\
                                 & 32   & $16\times3$        & 0.066                   &          0.077  & 301           \\
                                 & 32   & $32\times3$        & 0.062                   &          0.070  & 333           \\
                                 & 32   & $64\times3$        & 0.019                   &          0.057  & 312          \\
    \bottomrule
\end{tabular}
    \label{tab:results_klein_gordon}
\end{table}

The performance of the proposed approaches in benchmarking experiments is presented in \Cref{tab:results_klein_gordon}.
The results indicate that CP-PINN, TT-PINN, and Tucker-PINN demonstrate computational efficiency, requiring fewer collocation points and achieving higher accuracy (a factor of 10) compared to current state-of-the-art methods.
A comparative analysis of CP-PINN, TT-PINN, and Tucker-PINN reveals that increasing the number of collocation points and the model's rank generally leads to improved solution accuracy.
In particular, for both PDE problems, we observe that CP-PINN with a rank of $32$ and $64 \times 3$ collocation points achieves the most accurate solutions.
Similarly, TT-PINN achieves the second-best performance with the same rank and collocation points.
We hypothesize that a low rank may result in excessive information compression, potentially leading to losing essential details in the subsequent inverse decomposition.
\begin{table}[t]
\centering
    \caption{
        We show the $\mathcal{L}_2$ loss for the CP-PINN, TT-PINN, and Tucker-PINN on the 5d Poisson's Equation and (2+1)d Flow mixing simulation data.
        The best performing scores are \underline{underlined}, and each experiment has been repeated ten times (we omit the standard deviation as it has been consistent around \emph{0.01}.)
    }
    \begin{tabular}{lp{1.5cm}<{\centering}cp{1.6cm}<{\centering}p{1.7cm}<{\centering}c}
    \toprule
    Model & Points & Rank & CP-PINN & TT-PINN & Tucker-PINN \\
    \midrule
    \multirow[c]{3}{*}{5d Poisson's Equation} &\multirow[c]{3}{*}{$24 \times 3$} &  6 & 0.097 & 0.048 & \underline{0.040} \\
                                              &                                  &  8 & 0.077 & 0.047 & \underline{0.038} \\
                                              &                                  & 12 & 0.033 & 0.037 & \underline{0.026} \\
    \midrule
    (2+1)d Flow mixing                        &   $128\times3$                   & 64 & \underline{0.013} & 0.018 & 0.028 \\ 
    \bottomrule
    \end{tabular}
    \label{tab:5dpoisson}
\end{table}

To further substantiate our findings, we investigate whether our \emph{function tensor decomposition} scales to a higher dimension ($>3$) effectively, as shown theoretically.
Therefore, we solve 5d Poisson's Equation~\cite{zeng2024feature} and simulate the (2+1)d flow mixing PDE~\cite{CHIU2022114909} to capture the intricate mixing process of two fluids, see \Cref{tab:list_of_pdes} lower half.
The other PINN architectures are not suitable to solve this task owing to their vast sampling of collocation points and slow speed and are omitted from the results in \Cref{tab:5dpoisson}.
For the experiments, we use a fixed collocation point amount but vary the rank of the decomposition components.
Surprisingly, our experiments reveal that Tucker-PINN outperforms both CP-PINN and TT-PINN, contradicting the findings in \Cref{tab:results_klein_gordon}.
We attribute this discrepancy to the fact that CP decomposition has fewer parameters, which may make it more challenging to find suitable rank-one approximations as dimensionality increases.
Further, we observe that the (2+1)d flow mixing problem was solved only with a rank of $64$ by our proposed architectures.
This suggests that more collocation points may be necessary for an accurate approximation.
Additionally, modifications to the neural network architecture or training protocol and extensive hyperparameter tuning could enable solutions for lower ranks than $64$ but are out of the scope of this work.

\section{Discussion and Conclusions}
We introduce \emph{functional tensor decomposition} based PINNs, a novel approach for solving PDEs using PINNs.
Our essential contribution is extending the classical variable separable method to PINNs by leveraging function tensor decomposition forms.  
We show that PINNs can approximate multivariate PDEs by decomposing the solution as the outer product of smaller tensors with controlled ranks, enabling efficient and effective solutions to complex problems.

A critical insight is that such a PINN can learn any PDE irrespective of whether the variable separable form exists.
We investigate three tensor decomposition forms incorporated into the general PINN framework to reduce the computational complexity, especially the \emph{curse of dimensionaly}. 
The primary concept underlying our functional tensor decomposition approach for PINNs is as follows:
(1) We decompose the multivariate solution of a PDE and learn each part using an individual neural network, and (2) we use this within the PINN framework, where a loss function with PDE residual and boundary conditions is optimized.
Please note that a unique form of decomposition does not need to exist as we estimate the decomposition in a data-driven manner. 

We conducted experiments using benchmark PDEs, such as the Klein-Gordon Equation~\cite{cho2023separable} and 3d Helmholtz~\cite{takamoto2023pdebench} equation, to demonstrate that our proposed methods significantly outperform existing PINNs.
This distinction in accuracy is evident even when employing low numbers of collocation points and ranks.
Furthermore, we substantiated our findings by checking whether our method scales to higher dimensional ($> 3$) PDEs, such as the 5d Poisson equation~\cite{zeng2024feature} or (2+1)d flow mixing~\cite{CHIU2022114909}.
We are demonstrating the ability of our method to tackle complex issues with a relatively small number of collocation points and faster speed.
The results show that we effectively mitigate the \emph{curse of dimensionality}, enabling PINNs to solve problems in even higher dimensions efficiently.

Several open questions remain unanswered and could provide further interesting research in developing PINNs for high-dimensional PDEs, even though we already outperform existing methods by a factor of ten.
Our experiments indicate no best tensor decomposition form, and factors like collocation points, rank, and available physical knowledge play a significant role in overall performance.
The Canonical-Polyadic decomposition~\cite{cphitchcock} is the most straightforward representation among the same rank but becomes unstable in higher ranks~\cite{ttdecomp}.
While Tucker decomposition~\cite{Tucker1966}, with its increased parameter count for the same rank, does experience a curse of dimensionality due to the core tensor, albeit to a lesser extent than others.
We assume that Tensor-Train~\cite{ttdecomp} is a good candidate with lower dimensionality and is more stable in higher dimensions. We demonstrate the potential of \emph{functional tensor decomposition} in enhancing PINNs.
While further optimization may be possible, our findings already highlight the benefits of this method over traditional numerical approaches.

\bibliographystyle{splncs04}
\bibliography{paper}

\begin{thebibliography}{10}
\providecommand{\url}[1]{\texttt{#1}}
\providecommand{\urlprefix}{URL }
\providecommand{\doi}[1]{https://doi.org/#1}

\bibitem{Berkhahn2022}
Berkhahn, S., Ehrhardt, M.: A physics-informed neural network to model covid-19 infection and hospitalization scenarios. Advances in Continuous and Discrete Models  \textbf{2022}(1), ~61 (Oct 2022). \doi{10.1186/s13662-022-03733-5}

\bibitem{jax2018github}
Bradbury, J., Frostig, R., Hawkins, P., Johnson, M.J., Leary, C., Maclaurin, D., Necula, G., Paszke, A., Vander{P}las, J., Wanderman-{M}ilne, S., Zhang, Q.: {JAX}: composable transformations of {P}ython+{N}um{P}y programs (2018), \url{http://github.com/google/jax}

\bibitem{Cai2021}
Cai, S., Wang, Z., Wang, S., Perdikaris, P., Karniadakis, G.E.: {Physics-Informed Neural Networks for Heat Transfer Problems}. Journal of Heat Transfer  \textbf{143}(6),  060801 (04 2021). \doi{10.1115/1.4050542}

\bibitem{CHIU2022114909}
Chiu, P.H., Wong, J.C., Ooi, C., Dao, M.H., Ong, Y.S.: Can-pinn: A fast physics-informed neural network based on coupled-automatic–numerical differentiation method. Computer Methods in Applied Mechanics and Engineering  \textbf{395},  114909 (2022). \doi{https://doi.org/10.1016/j.cma.2022.114909}

\bibitem{cho2023separable}
Cho, J., Nam, S., Yang, H., Yun, S.B., Hong, Y., Park, E.: Separable physics-informed neural networks. Advances in Neural Information Processing Systems  (2023)

\bibitem{coscia2023physics}
Coscia, D., Ivagnes, A., Demo, N., Rozza, G.: Physics-informed neural networks for advanced modeling. Journal of Open Source Software  \textbf{8}(87), ~5352 (2023)

\bibitem{Cuomo2022}
Cuomo, S., Di~Cola, V.S., Giampaolo, F., Rozza, G., Raissi, M., Piccialli, F.: Scientific machine learning through physics--informed neural networks: Where we are and what's next. Journal of Scientific Computing  \textbf{92}(3), ~88 (Jul 2022). \doi{10.1007/s10915-022-01939-z}

\bibitem{Cybenko1989}
Cybenko, G.: Approximation by superpositions of a sigmoidal function. Mathematics of Control, Signals and Systems  \textbf{2}(4),  303--314 (Dec 1989). \doi{10.1007/BF02551274}, \url{https://doi.org/10.1007/BF02551274}

\bibitem{glorot10a}
Glorot, X., Bengio, Y.: Understanding the difficulty of training deep feedforward neural networks. In: Teh, Y.W., Titterington, M. (eds.) Proceedings of the Thirteenth International Conference on Artificial Intelligence and Statistics. Proceedings of Machine Learning Research, vol.~9, pp. 249--256. PMLR, Chia Laguna Resort, Sardinia, Italy (13--15 May 2010), \url{https://proceedings.mlr.press/v9/glorot10a.html}

\bibitem{HAGHIGHAT2021113741}
Haghighat, E., Raissi, M., Moure, A., Gomez, H., Juanes, R.: A physics-informed deep learning framework for inversion and surrogate modeling in solid mechanics. Computer Methods in Applied Mechanics and Engineering  \textbf{379},  113741 (2021). \doi{https://doi.org/10.1016/j.cma.2021.113741}

\bibitem{phdthesis}
Herath, I.: Multivariate Regression using Neural Networks and Sums of Separable Functions. Ph.D. thesis, Ohio University (04 2022), \url{http://rave.ohiolink.edu/etdc/view?acc_num=ohiou1648166101093853}

\bibitem{cphitchcock}
Hitchcock, F.L.: The expression of a tensor or a polyadic as a sum of products. Journal of Mathematics and Physics  \textbf{6}(1-4),  164--189 (1927). \doi{https://doi.org/10.1002/sapm192761164}

\bibitem{HORNIK1989359}
Hornik, K., Stinchcombe, M., White, H.: Multilayer feedforward networks are universal approximators. Neural Networks  \textbf{2}(5),  359--366 (1989). \doi{https://doi.org/10.1016/0893-6080(89)90020-8}

\bibitem{JinMIO}
Jin, P., Meng, S., Lu, L.: Mionet: Learning multiple-input operators via tensor product. SIAM Journal on Scientific Computing  \textbf{44}(6),  A3490--A3514 (2022). \doi{10.1137/22M1477751}

\bibitem{kingma2014adam}
Kingma, D.P., Ba, J.: Adam: A method for stochastic optimization. arXiv preprint arXiv:1412.6980  (2014)

\bibitem{TensorDecompsReview}
Kolda, T.G., Bader, B.W.: Tensor decompositions and applications. SIAM Review  \textbf{51}(3),  455--500 (2009). \doi{10.1137/07070111X}

\bibitem{krishnapriyan2021characterizing}
Krishnapriyan, A.S., Gholami, A., Zhe, S., Kirby, R.M., Mahoney, M.W.: Characterizing possible failure modes in physics-informed neural networks (2021)

\bibitem{Lin_Maxey_Li_Karniadakis_2021}
Lin, C., Maxey, M., Li, Z., Karniadakis, G.E.: A seamless multiscale operator neural network for inferring bubble dynamics. Journal of Fluid Mechanics  \textbf{929}, ~A18 (2021). \doi{10.1017/jfm.2021.866}

\bibitem{Lu2021}
Lu, L., Jin, P., Pang, G., Zhang, Z., Karniadakis, G.E.: Learning nonlinear operators via deeponet based on the universal approximation theorem of operators. Nature Machine Intelligence  \textbf{3}(3),  218--229 (Mar 2021). \doi{10.1038/s42256-021-00302-5}

\bibitem{LuLu2019}
Lu, L., Meng, X., Mao, Z., Karniadakis, G.E.: Deepxde: A deep learning library for solving differential equations (7 2019). \doi{10.1137/19M1274067}

\bibitem{lu2021deepxde}
Lu, L., Meng, X., Mao, Z., Karniadakis, G.E.: {DeepXDE}: A deep learning library for solving differential equations. SIAM Review  \textbf{63}(1),  208--228 (2021). \doi{10.1137/19M1274067}

\bibitem{Maddu_2022}
Maddu, S., Sturm, D., Müller, C.L., Sbalzarini, I.F.: Inverse dirichlet weighting enables reliable training of physics informed neural networks. Machine Learning: Science and Technology  \textbf{3}(1),  015026 (feb 2022). \doi{10.1088/2632-2153/ac3712}

\bibitem{mcclenny2022selfadaptive}
McClenny, L., Braga-Neto, U.: Self-adaptive physics-informed neural networks using a soft attention mechanism (2022)

\bibitem{moseley2021finite}
Moseley, B., Markham, A., Nissen-Meyer, T.: Finite basis physics-informed neural networks (fbpinns): a scalable domain decomposition approach for solving differential equations (2021)

\bibitem{ttdecomp}
Oseledets, I.V.: Tensor-train decomposition. SIAM Journal on Scientific Computing  \textbf{33}(5),  2295--2317 (2011). \doi{10.1137/090752286}

\bibitem{raisinghaniaordinary}
Raisinghania, M.: Ordinary and Partial Differential Equations. S. Chand Publishing (1991), \url{https://books.google.de/books?id=vaorDAAAQBAJ}

\bibitem{Raissi2017}
Raissi, M., Perdikaris, P., Karniadakis, G.E.: Physics informed deep learning (part ii): Data-driven discovery of nonlinear partial differential equations (11 2017), \url{http://arxiv.org/abs/1711.10566}

\bibitem{rudin1964principles}
Rudin, W.: Principles of Mathematical Analysis. International series in pure and applied mathematics, McGraw-Hill (1964), \url{https://books.google.de/books?id=yifvAAAAMAAJ}

\bibitem{takamoto2023pdebench}
Takamoto, M., Praditia, T., Leiteritz, R., MacKinlay, D., Alesiani, F., Pflüger, D., Niepert, M.: Pdebench: An extensive benchmark for scientific machine learning (2023)

\bibitem{Tucker1966}
Tucker, L.R.: Some mathematical notes on three-mode factor analysis. Psychometrika  \textbf{31}(3),  279--311 (Sep 1966). \doi{10.1007/BF02289464}

\bibitem{soilsai}
Vemuri, S.K., B{\"u}chner, T., Denzler, J.: Estimating soil hydraulic parameters for unsaturated flow using physics-informed neural networks. In: Franco, L., de~Mulatier, C., Paszynski, M., Krzhizhanovskaya, V.V., Dongarra, J.J., Sloot, P.M.A. (eds.) Computational Science -- ICCS 2024. pp. 338--351. Springer Nature Switzerland, Cham (2024). \doi{10.1007/978-3-031-63759-9_37}

\bibitem{vemuri}
Vemuri, S.K., Denzler, J.: Gradient statistics-based multi-objective optimization in physics-informed neural networks. Sensors  \textbf{23}(21) (2023). \doi{10.3390/s23218665}

\bibitem{wang2022respecting}
Wang, S., Sankaran, S., Perdikaris, P.: Respecting causality is all you need for training physics-informed neural networks (2022)

\bibitem{Wangntk}
Wang, S., Teng, Y., Perdikaris, P.: Understanding and mitigating gradient flow pathologies in physics-informed neural networks. SIAM Journal on Scientific Computing  \textbf{43}(5),  A3055--A3081 (2021). \doi{10.1137/20M1318043}

\bibitem{WANG2022110768}
Wang, S., Yu, X., Perdikaris, P.: When and why pinns fail to train: A neural tangent kernel perspective. Journal of Computational Physics  \textbf{449},  110768 (2022). \doi{10.1016/j.jcp.2021.110768}

\bibitem{Yu_2022}
Yu, J., Lu, L., Meng, X., Karniadakis, G.E.: Gradient-enhanced physics-informed neural networks for forward and inverse pde problems. Computer Methods in Applied Mechanics and Engineering  \textbf{393},  114823 (Apr 2022). \doi{10.1016/j.cma.2022.114823}

\bibitem{zeng2024feature}
Zeng, C., Burghardt, T., Gambaruto, A.M.: Feature mapping in physics-informed neural networks (pinns) (2024)

\end{thebibliography}

\end{document}